\documentclass[10pt,twocolumn,letterpaper]{article}

\usepackage{cvpr}              

\usepackage[dvipsnames]{xcolor}

\definecolor{cvprblue}{rgb}{0.21,0.49,0.74}
\usepackage[pagebackref,breaklinks,colorlinks,citecolor=cvprblue]{hyperref}
\usepackage{booktabs}
\usepackage{multirow}
\usepackage{indentfirst}
\usepackage[ruled]{algorithm2e}
\usepackage{amsmath}
\usepackage{colortbl}

\SetKwComment{Comment}{/*}{*/}
\def\paperID{128} 
\def\confName{CVPR}
\def\confYear{2024}

\title{Domain-Specific Block Selection and Paired-View Pseudo-Labeling for \\Online Test-Time Adaptation}

\author{Yeonguk Yu, Sungho Shin, Seunghyeok Back, Minhwan Ko, Sangjun Noh, and Kyoobin Lee
\and
Gwangju Institute of Science and Technology\\
{\tt\small \{yeon\_guk, hogili89, shback, mhko1998, sangjun7\}@gm.gist.ac.kr, kyoobinlee@gist.ac.kr}
}

\begin{document}
\maketitle
\begin{abstract}

Test-time adaptation (TTA) aims to adapt a pre-trained model to a new test domain without access to source data after deployment. Existing approaches typically rely on self-training with pseudo-labels since ground-truth cannot be obtained from test data. Although the quality of pseudo labels is important for stable and accurate long-term adaptation, it has not been previously addressed. In this work, we propose DPLOT, a simple yet effective TTA framework that consists of two components: (1) domain-specific block selection and (2) pseudo-label generation using paired-view images. Specifically, we select blocks that involve domain-specific feature extraction and train these blocks by entropy minimization. After blocks are adjusted for current test domain, we generate pseudo-labels by averaging given test images and corresponding flipped counterparts. By simply using flip augmentation, we prevent a decrease in the quality of the pseudo-labels, which can be caused by the domain gap resulting from strong augmentation. Our experimental results demonstrate that DPLOT outperforms previous TTA methods in CIFAR10-C, CIFAR100-C, and ImageNet-C benchmarks, reducing error by up to 5.4\%, 9.1\%, and 2.9\%, respectively. Also, we provide an extensive analysis to demonstrate effectiveness of our framework. Code is available at \href{https://github.com/gist-ailab/domain-specific-block-selection-and-paired-view-pseudo-labeling-for-online-TTA}{https://github.com/gist-ailab/domain-specific-block-selection-and-paired-view-pseudo-labeling-for-online-TTA}.
\end{abstract}

\section{Introduction}
\label{sec:intro}

Deep neural networks achieve remarkable performance when the training and target data originate from the same domain \cite{SimonyanZ14a, renNIPS15fasterrcnn}. In contrast, deployed models perform poorly if domain shifts exists between source training data and target test data \cite{recht2019imagenet, geirhos2018generalisation}. For example, a pre-trained image classification model may suffer this phenomenon for the given corrupted images due to sensor degradation, weather change, and other reasons \cite{schneider2020improving, hendrycks2019robustness}. Various studies have addressed the domain shift issue \cite{hoyer2023mic, zhou2022domain, lu2022domain}. Recently, test-time adaptation (TTA), which aims to improve the model performance on target domain without access to the source data during the inference stage, has received attention because of its practicality and applicability \cite{wang2021tent, wang2022continual, niu2022efficient}.

\begin{figure}[t]
    \centering
    \includegraphics[width=0.49\textwidth]{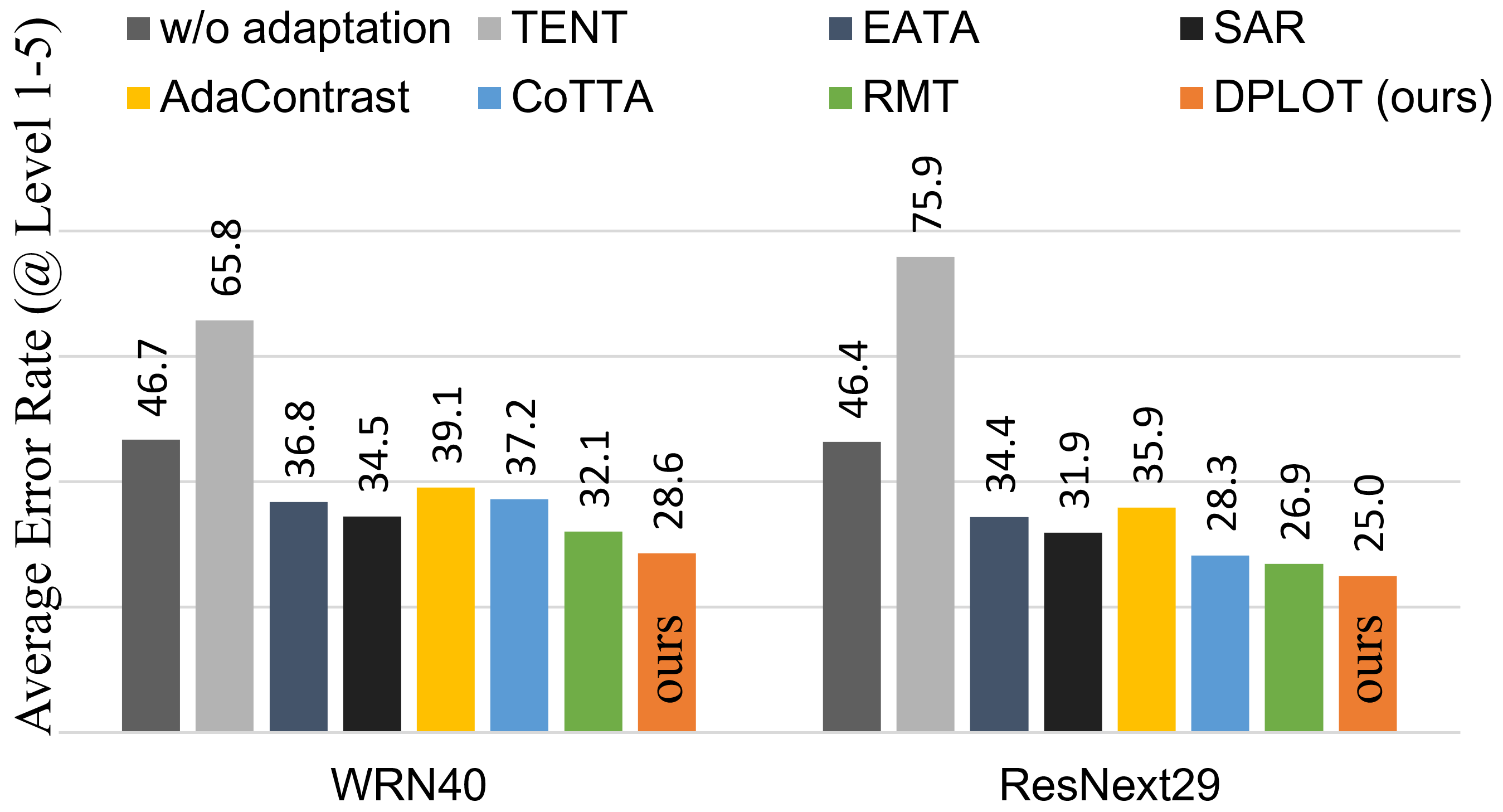}
    \caption{\textbf{Results of the proposed framework for online test-time adaptation (Orange).} We evaluate average error rates of the WideResNet40 and ResNext-29 architectures for the CIFAR100-C gradual setting benchmark using competitive test-time adaptation methods. In the gradual setting, the networks should adapt to continually changing corruption domains (135 changes in total).}
    \label{fig:overview}
\end{figure}

In online TTA, the objective is to simultaneously make prediction and adaptation using a source pre-trained model for the given test data. Existing TTA methods typically rely on self-training with pseudo labels, such as entropy minimization \cite{grandvalet2004semi, vu2019advent} and consistency regularization \cite{tarvainen2017mean, sohn2020fixmatch}. Entropy minimization trains the model by self-generated pseudo-labels, whereas consistency regularization uses pseudo-labels generated by a teacher model. These methods have demonstrated excellent performance with short-term test sequence in the stationary environment~\cite{wang2021tent}.

Under the continually changing domain in non-stationary environments \cite{marsden2022gradual, wang2022continual}, self-training with pseudo-labels can lead to error accumulation, gradually degrading the quality of pseudo-labels \cite{wang2022continual}. To alleviate the issue, stochastic restoration (i.e., reset the model to the source pre-trained weight) and augmentation-averaged pseudo-label (i.e., various hard augmentations including color jitter) was proposed by \cite{wang2022continual}. In addition, D{\"o}bler \etal~\cite{dobler2023robust} combined symmetric cross-entropy \cite{wang2019sce} and contrastive loss for stabilizing the pseudo-labeling without restoration. However, previous methods lacks an accurate approach to generate pseudo-labels from the teacher, a critical aspect for stable long-term test-time adaptation.

In this study, we address accurate pseudo-label generation in underlying assumption that \textit{images from source domain and target domain share domain-invariant features for a task, regardless of shifted domain-specific features caused by corruptions}. We consider the domain-invariant feature to be a high-level feature useful for the task and the domain-specific feature to be a low-level feature with no concern for the task as termed in \cite{wang2021domain, wu2021vector, wu2021instance}. To achieve precise adaptation under this assumption, we propose \textbf{D}omain-specific block selection and paired-view \textbf{P}seudo-\textbf{L}abeling for \textbf{O}nline \textbf{T}est-Time adaptation (DPLOT), which is composed of two core components: (1) domain-specific block selection and (2) pseudo-label generation using paired-view images. The block selection method identifies blocks involve in domain-specific feature extraction (termed domain-specific block) using augmented source training data. These domain-specific blocks are then fine-tuned through entropy minimization during the test-time phase. Then, we generate the pseudo-label from the teacher by averaging predictions for the given test image and its horizontally flipped counterpart to update all model parameters. This is motivated by the fact that the teacher model's domain-specific blocks are adjusted for the current domain, and hard augmentation may generate another domain gap, which may lead to degraded pseudo-labels. Consequently, our framework provides strong adaptation performance for the model as shown in Figure \ref{fig:overview}.

In summary, the main contributions are as follows:
\begin{itemize}
    \item We propose the DPLOT, which consists of domain-specific block selection and paired-view pseudo labeling for long-term online test-time adaptation.
    \item We compare the proposed method with other online TTA methods in both \textit{continual} and \textit{gradual} settings benchmarks, and our framework outperforms other methods. 
    \item We provide a wide range of analyses that lead to an improved understanding of our framework.
\end{itemize}

\section{Related Work}
\label{sec:related_work}
\paragraph{Unsupervised Domain Adaptation}
Unsupervised domain adaptation (UDA) aims to adapt a model to a target domain using given labeled source data and unlabeled target data before model deployment.For example, Ganin and Lempitsky~\cite{ganin2015unsupervised} proposed gradient reversal layer to force the feature extractor to produce same feature distribution for the given source data and target data . Also, Hoffman~\etal~\cite{hoffman2018cycada} used image-to-image translation to create labeled target-like source data to train the network. French~\etal~\cite{french2018selfensembling} used self-ensembling with the mean teacher to minimize the difference between the student's prediction for the augmented test data and the teacher's prediction for the test data. Kundu~\etal\cite{kundu2020class} proposed class-incremental method without using source training data. Recently, Hoyer~\etal~\cite{hoyer2023mic} proposed the adaptation framework based on masked image consistency, where the model is forced to produce the same prediction for the given target data and corresponding masked data. Also, Prasanna~\etal\cite{prasanna2023continual} proposed the continual domain adaptation method based on pruning-aided weight modulation to reduce catastrophic forgetting. Since both UDA and TTA aim to performance improvement in target domain, the methods from UDA can be considered. Our method also uses the mean teacher as in \cite{hoyer2023mic} to reduce the prediction difference for adapting the network to the target domain but without using source data.

\paragraph{Test-Time Adaptation}
Test-time adaptation (TTA), which only requires target data for adaptation unlike UDA, has gained increasing attention. As in \cite{liang2023ttasurvey}, (online) TTA can be categorized into batch normalization (BN) calibration \cite{mirza2022norm, zhao2023delta}, entropy minimization \cite{wang2021tent, niu2022efficient, niu2023towards, choi2022improving}, and consistency regularization \cite{wang2022continual, dobler2023robust, wang2022contrastive}. Specifically, Mizra~\etal~\cite{mirza2022norm} proposed dynamic unsupervised adaptation (DUA), which adapts the statistics of the BN layers\cite{ioffe2015batch} to remove degradation caused by BN\cite{li2017revisiting, galloway2019batch}. On the other hand, updating affine parameters of BN layers has demonstrated improved adaptation performance \cite{wang2021tent, niu2022efficient, niu2023towards, choi2022improving}. Niu~\etal~\cite{niu2023towards} demonstrated that using batch-independent normalization methods like layer norm \cite{ba2016layer} and group norm \cite{wu2018group} instead of using batch-dependent BN is helpful for stable entropy minimization. Also, they proposed sharpness-aware entropy minimization, which filters out unreliable samples by their gradients, based on the observation that large gradient samples lead to model collapse. Similar to our work, Choi~\etal~\cite{choi2022improving} proposed the framework that updates model parameters by entropy minimization, differently depending on their sensitivity of distribution shift. Simultaneously, consistency regularization-based TTA frameworks have been investigated. Wang~\etal~\cite{wang2022continual} proposed TTA in a continually changing domain by using augmentation-averaged pseudo-target from the mean teacher. Also, Chen~\etal~\cite{chen2022contrastive} proposed to use contrastive learning in TTA for learning better representation in the target domain. Moreover, robust mean teacher~(RMT) is proposed \cite{dobler2023robust}, which showed state-of-the-art performance by using various techniques such as symmetric cross-entropy and source replay. Previous entropy minimization methods \cite{wang2021tent, niu2022efficient, niu2023towards} update BN layer of the model which can modify domain-invariant feature extraction. In this work, we propose block selection method to prevent the modification of domain-invariant feature extraction by updating blocks that involve in domain-specific feature extraction. Also, we propose to generate pseudo-labels from the teacher only using simple flip augmentation to improve the quality of labels. Consequently, our DPLOT uses both entropy minimization and consistency regularization for reliable long-term adaptation.

\section{Method}
\label{sec:method}
We introduces DPLOT, a simple yet effective online TTA framework. First, we describe the overview of our framework in Section \ref{subsec:motivation}. Second, we describe the block selection method before deployment in Section \ref{subsec:block}. Lastly, we introduce how our framework adapts the model for the given unlabeled test data after deployment in Section \ref{subsec:pcloss}.

\subsection{Overview of DPLOT's components}
\label{subsec:motivation}

\begin{figure}[t]
    \centering
    \includegraphics[width=0.49\textwidth]{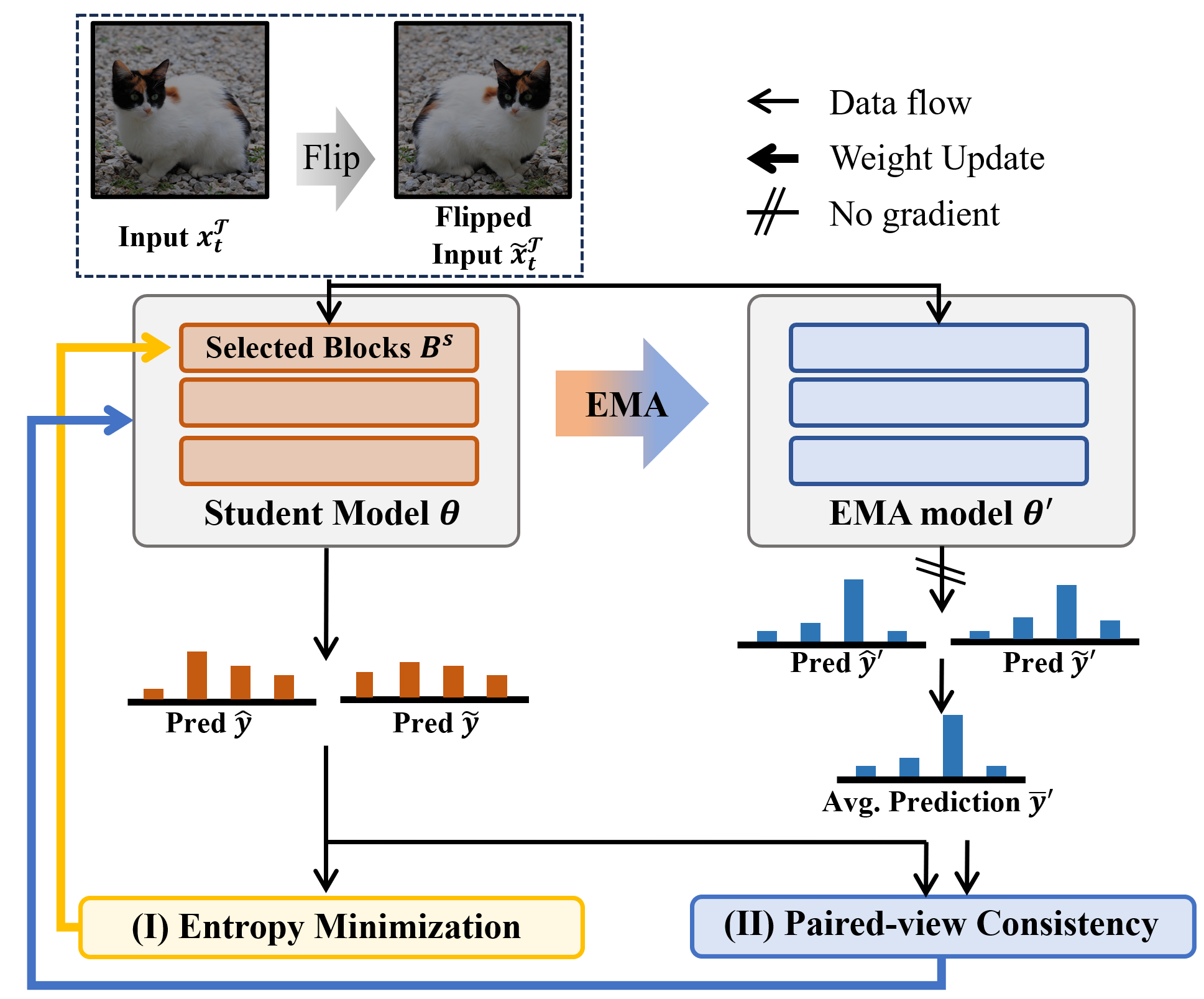}
    \caption{Illustration of our proposed test-time adaptation using entropy minimization and paired-view consistency. During test-time, the current test and corresponding flipped images are given to the student model and EMA teacher model. Entropy minimization is performed to update the parameters of selected blocks (yellow arrow), while all parameters are updated to minimize the difference between student output and the averaged EMA model's prediction (blue arrow).}
    \label{fig:framework}
\end{figure}

The components within our framework can be categorized based on whether they are used before or after deployment. Domain-specific block selection is conducted before deployment, while our adaptation method as shown in Figure \ref{fig:framework}, involving entropy minimization on the selected blocks and the use of paired-view pseudo-labels for updating all model parameters, is performed after deployment.

The domain-specific block selection is proposed under the assumption that the domain-invariant feature of test image remains consistent with the source domain, while the domain-specific feature changes. Therefore, the domain-specific feature extraction of the model should be adjusted to bridge the gap between the source domain and target domain caused by different feature statistics \cite{li2017revisiting, benz2021revisiting, wang2021domain}. We evaluate each block by measuring cosine similarity between prototype features before and after the entropy minimization using Gaussian noise-added source data. Subsequently, we select the blocks that maintain high similarity, indicating that the blocks do not involve in domain-invariant feature extraction (i.e., involves in domain-specific feature extraction). During test-time, the selected blocks are updated by minimizing entropy for the given test data to adjust the domain-specific feature extraction to the current corruption. By selecting domain-specific blocks, we can adjust domain-specific feature extraction without disrupting domain-invariant feature extraction unlike previous method \cite{wang2021tent}, where all batch normalization \cite{ioffe2015batch} layers are updated.

We provide pseudo-labels generated by the exponential moving average (EMA) teacher \cite{tarvainen2017mean} (i.e., $\theta^\prime_{t+1} = \alpha \theta^\prime_t + (1-\alpha)\theta_{t+1}$) to the model for further adjusting all parameters on the target domain. This is based on insight that the high-level feature in the test image can be affected by domain shifts, and all parameters should be adjusted to increase adaptation performance. In contrast to previous approach \cite{wang2022continual}, where hard augmentations (e.g., color jitter, Gaussian noise, blur, and random pad-crop) are used for generating pseudo-labels, we only use horizontal flip augmentation. Since the teacher's domain-specific extraction is adjusted on the target domain by entropy minimization and even a small domain gap between the test and augmented images can reduce the teacher's accuracy, the horizontal flip is well suited for generating pseudo-labels from the teacher.



\subsection{Block selection before deployment}
\label{subsec:block}
We consider a pre-trained neural network $\theta$, comprising a feature extractor and a classifier. We assume that the feature extractor of the network is composed of multiple $L$ blocks (e.g., ResNet18 has 8 residual blocks \cite{he2016deep}). For a given a RGB image $x\in \mathbb{R}^{H\times W\times 3}$, a feature vector $f\in\mathbb{R}^d$ of dimension $d$ is extracted. Then, our objective is to select blocks that involve in domain-specific feature extraction. To this end, we first calculate the domain-invariant feature space using prototype vectors. The prototype vector $p_c\in\mathbb{R}^d$ is acquired by averaging all source feature vectors belonging to the class $c$ as follows:
\begin{equation}
    p_c = \frac{1}{N_c}\sum_{n = 1}^{N_c} f(x^c_n\in \mathcal{X}^S;\theta),
\label{eq:1}
\end{equation}
where $f(\cdot), x^c_n$, and $N_c$ refer to the feature extraction, $n$-th RGB image belonging to the class $c$, and number of images for class $c$, respectively. Consequently, prototypes for all class, $P=\{p_1, p_2, ..., p_C\}$, are obtained.

To further select the block for adjusting domain-specific feature extraction during test-time, we measure the similarity $s$ between the original prototypes and modified prototypes after the block's parameters have been updated via entropy minimization with Gaussian noise (zero mean and 0.5 variance) added training images. We use Gaussian noise as it is common corruption and can represent various domain shifts \cite{lopes2019improving, cygert2020toward}. This similarity is calculated by averaging cosine similarity as follows:
\begin{equation} 
    s_i = \frac{1}{C}\sum_{c = 1}^C \frac{p_c \cdot p^\prime_c}{\|p_c\|\|p^\prime_c\|},
\label{eq:2}
\end{equation}
where $p^\prime_c$ denotes a single modified prototype vector for the $c$-th class after the entropy minimization. Also, $C$ refers to the number of classes. After computing the similarity for every block, we apply min-max scaling to normalize the results into the range of [0, 1]. This scaling allows us to choose threshold more generally across various architectures. A high similarity indicates that the block is adapted on Gaussian noise added images without modifying domain-invariant feature space (i.e., high-level feature space). Finally, we select blocks that higher than the threshold $\gamma$. The pseudo code of the block selection is described in Algorithm~\ref{alg:bs}. In our experiments, the threshold is set to 0.75, unless otherwise specified.

\begin{algorithm}[t]
\caption{Domain-specific block selection}
\label{alg:bs}
\SetKwInOut{Input}{Input}
\SetKwInOut{Output}{Output}
\SetKwInOut{Initialization}{Init.}

\Initialization{$\text{An empty list } \mathcal{S}, \text{an empty list } \mathcal{B}^s;$}
\Input{Blocks of the model $\mathcal{B}=\{b_1, b_2, ..., b_L\}$, source data $(\mathcal{X}^S, \mathcal{Y}^S)$, threshold $\gamma$, pre-trained model $\theta$;}

Calculate $P$ using Eq. \ref{eq:1}\\ 
Add noise to $\mathcal{X}^S$\\

\ForEach{block $b_i \in \mathcal{B}$}{    
    Minimize entropy for parameter of $b_i$ using $\mathcal{X}^S$\\

    Calculate $P^\prime$ using Eq. \ref{eq:1}\\
    Calculate $s_i$ using Eq. \ref{eq:2}\\
    Append $s_i$ in $\mathcal{S}$\\
    Reset $\theta$;
}
Min-max scale $S$ \\
\ForEach{$(s_i, b_i) \in (\mathcal{S}, \mathcal{B})$}{
\If{$s_i>\gamma$}{
    Append $b_i$ $\text{in}$ $\mathcal{B}^s;$
    }
}
\Output{Selected block list $\mathcal{B}^s$}
\end{algorithm}

\begin{algorithm}[t]
\caption{Adaptation process after deployment}
\label{alg:tta}
\SetKwInOut{Input}{Input}
\SetKwInOut{Output}{Output}
\SetKwInOut{Initialization}{Init.}

\Initialization{Selected Blocks $\mathcal{B}^s$, a pre-trained model $\theta$, a teacher model $\theta^\prime$ initialized from $\theta$;}
\Input{For each time step $t$, current batch of data~$x_t^\mathcal{T}$;}
1: Horizontally flip $x_t^\mathcal{T}$ and get pseudo-label from $\theta^\prime_t$ by Eq.~\ref{eq:pred} and Eq.~\ref{eq:pred_flipped}\\
2: Update $\mathcal{B}^s$ of $\theta_t$ by entropy minimization loss $\mathcal{L}_e$ in Eq.~\ref{eq:em} using both $x_t^\mathcal{T}$ and $\tilde{x}_t^\mathcal{T}$\\
3: Update $\theta_t$ by paired-view consistency loss $\mathcal{L}_pc$ in Eq.~\ref{eq:pc} using both $x_t^\mathcal{T}$ and $\tilde{x}_t^\mathcal{T}$\\
4: Update $\theta^\prime_t$ by moving average of $\theta_t$\\

\Output{Ensemble prediction, Updated model $\theta_{t+1}$, Updated teacher $\theta^\prime_{t+1}$}
\end{algorithm}

\subsection{Test-time adaptation after deployment}
\label{subsec:pcloss}
\paragraph{Entropy minimization}
Entropy minimization is performed for the given target data at current time $x_t^T$ to update the selected blocks $\mathcal{B}^s$ of model $\theta$. Following other entropy minimization-based methods \cite{wang2021tent, niu2022efficient, niu2023towards, choi2022improving}, we use Shannon Entropy \cite{6773024} as follows:
\begin{equation}
\label{eq:em}
    \mathcal{L}_e = -\sum_c \hat{y}_c \log \hat{y}_c,
\end{equation}
where $\hat{y}_c$ represents the output probability for the $c$-th class using the model $\theta$. By minimizing entropy on test batch $x^\mathcal{T}_t$, the model is trained to push decision boundaries toward low-density region in prediction space \cite{vu2019advent}. As a result, the model is forced to acquire discriminative high-level features from current target domain by adjusting domain-specific feature extraction.

\paragraph{Paired-view consistency}
Domain-specific feature extraction of the model is adjusted to the current corruption using entropy minimization with selected domain-specific blocks. To leverage all parameters, the consistency regularization between the model $\theta$ and the moving average teacher $\theta^\prime$, which makes the training stable \cite{tarvainen2017mean, wang2022continual, dobler2023robust}, is used. Specifically, we use horizontal flip and moving average teacher \cite{tarvainen2017mean} to generate pseudo-labels for the given test and corresponding flipped images as follows:
\begin{equation}
    \label{eq:pred}
    \hat{y}^\prime = g_{\theta^\prime}(x_t^\mathcal{T}),
\end{equation}
\begin{equation}
    \label{eq:pred_flipped}
    \tilde{y}^\prime = g_{\theta^\prime}(\tilde{x}_t^\mathcal{T}),
\end{equation}
where $g_{\theta^\prime}(\cdot)$ refers to the teacher's prediction of the given input. Also, $x$ and $\tilde{x}$ refer to the original input and horizontally flipped input. The pseudo-label $\bar{y}^\prime$ is calculated by averaging $\hat{y}^\prime$ and $\tilde{y}^\prime$. Subsequently, all parameters of the model $\theta$ are updated by the symmetric cross-entropy \cite{dobler2023robust, wang2019sce} between the model and teacher predictions:
\begin{equation}
\label{eq:pc}
    \mathcal{L}_{pc} = \mathcal{L}_{sce}(\hat{y}, \bar{y}^\prime) + \mathcal{L}_{sce}(\tilde{y}, \bar{y}^\prime),
\end{equation}
\begin{equation}
    \mathcal{L}_{sce}(a, b) = \frac{1}{2}\cdot(\mathcal{L}_{ce}(a, b) + \mathcal{L}_{ce}(b, a)),
\end{equation}
where $\mathcal{L}_{sce}$ and $\tilde{y}$ refer to the symmetric cross-entropy and the model's prediction for the flipped input, respectively. Also, $L_{ce}$ refers to the standard cross-entropy loss. We use the symmetric cross-entropy since it is known to be robust to noisy labels \cite{dobler2023robust, wang2019sce}. After adaptation, we ensemble the predictions of both models by adding the model’s and teacher’s logits as \cite{dobler2023robust} for better performance. Our method during test-time is summarized in Algorithm \ref{alg:tta}.

\section{Experiments}
\label{sec:experiment}

\paragraph{Setup}
We evaluate our framework on CIFAR10-C, CIFAR100-C, and ImageNet-C, designed to benchmark the robustness of classification networks \cite{hendrycks2019robustness}. CIFAR dataset contains 10,000 and ImageNet dataset contains 50,000 test images for each of the 15 corruptions with 5 levels of severity. For the experiments, we use a network pre-trained on the clean training set of CIFAR \cite{krizhevsky2009learning} and ImageNet \cite{deng2009imagenet}. For CIFAR10, we use WRN28-10 \cite{Zagoruyko2016WRN}, WRN40-2A \cite{hendrycks2020augmix}, ResNet-18A \cite{kireev2022effectiveness}. For CIFAR100, we use WRN40-2A \cite{hendrycks2020augmix} and ResNext-29 \cite{xie2017aggregated}. For ImageNet, we use ResNet50 \cite{he2016deep} and ResNet50A \cite{hendrycks2021many}. Architectures named with~'A'~refers to the networks trained to be robust against corruption (e.g., AugMix \cite{hendrycks2020augmix}); specific details are described in supplementary materials.

We evaluate our method in two different settings. First, we consider the continual setting introduced by~\cite{wang2022continual}. Unlike the basic TTA setting, where evaluation is conducted for each corruption individually, the model is adapted to a sequence of test domains in an online fashion under the largest corruption severity level 5 (total 15 shifts). Second, we consider the gradual setting, as introduced by~\cite{marsden2022gradual} where the corruption severity level changes as follows: 1 $\rightarrow$ 2 $\rightarrow \cdots \rightarrow 5 \rightarrow \cdots \rightarrow$ 2 $\rightarrow$ 1 (total 135 shifts). This setting is motivated by the fact that domain shift does not occur abruptly but changes rather smoothly. 

Also, we follow the implementation setting of RMT~\cite{dobler2023robust}. Specifically, the batch size during test-time is set to 200 and 64 for CIFAR and ImageNet-C, respectively. We use an Adam~\cite{Kingma2014AdamAM} optimizer with a learning rate of 1e-3 and 1e-4 for entropy minimization to domain-specific blocks and paired-view consistency to all blocks, respectively. Warm-up is conducted before deployment, as done in \cite{dobler2023robust}, and we use pre-trained weights provided by RobustBench~\cite{croce2020robustbench}.

\paragraph{Baselines}

We compare our method with various source-free TTA baselines. Also, BN-1 refers to method that recalculates the BN statistics using the test batch. TENT~\cite{wang2021tent}, EATA~\cite{niu2022efficient}, and SAR~\cite{niu2023towards} are entropy minimization-based methods that update batch normalization layer weights to minimize the entropy of current predictions. AdaContrast~\cite{chen2022contrastive} relies on contrastive learning principles, combining contrastive learning and pseudo-labeling to enable discriminative feature learning for TTA. CoTTA\cite{wang2022continual} uses a teacher's augmentation-averaged pseudo-label for training and stochastic restore to mitigate error accumulation. RMT~\cite{dobler2023robust} uses the symmetric cross-entropy, which reduces effect of noisy label, for consistency regularization with pseudo-labels generated by a teacher and contrastive learning to pull the test feature space closer to the source domain. Moreover, RMT utilizes source replays during test-time to keep source knowledge. However, it is worth noting that we do not use source replays as it is not source-free during test-time. 

\begin{table}[t]
\centering
\huge
\resizebox{0.45\textwidth}{!}{
\begin{tabular}{@{}lcccc@{}}
\toprule
\multicolumn{1}{c}{\multirow{2}{*}{Method}} & \multicolumn{2}{c}{CIFAR100-C}                     & \multicolumn{2}{c}{ImageNet-C} \\ \cmidrule(l){2-5} 
\multicolumn{1}{c}{}                        & ResNext-29A    & \multicolumn{1}{c|}{WRN40-2A}      & ResNet-50      & ResNet-50A    \\ \midrule
\multicolumn{1}{l|}{Source only}            & 46.4†         & \multicolumn{1}{c|}{45.4}          & 82.0†           & 67.5          \\
\multicolumn{1}{l|}{BN-1}                   & 35.4†         & \multicolumn{1}{c|}{39.3}          & 68.6†           & 53.8          \\
\multicolumn{1}{l|}{TENT-cont.}             & 60.9†         & \multicolumn{1}{c|}{37.5}          & 62.6†           & 49.6          \\
\multicolumn{1}{l|}{AdaContrast}            & 33.4†         & \multicolumn{1}{c|}{37.1}          & 65.5†           & 50.9          \\
\multicolumn{1}{l|}{CoTTA}                  & 32.5†         & \multicolumn{1}{c|}{38.2}          & 62.7†           & 47.8          \\
\multicolumn{1}{l|}{EATA}                   & 32.3          & \multicolumn{1}{c|}{35.7}          & \textbf{58.8}  & 46.3          \\
\multicolumn{1}{l|}{SAR}                    & 31.9          & \multicolumn{1}{c|}{35.3}          & 61.9           & 49.3           \\
\multicolumn{1}{l|}{RMT}                    & 29.0†         & \multicolumn{1}{c|}{34.3}          & 59.8†           & 46.9          \\
\multicolumn{1}{l|}{DPLOT (ours)}            & \textbf{27.8} & \multicolumn{1}{c|}{\textbf{31.8}} & 60.2  & \textbf{44.6} \\ \bottomrule
\end{tabular}}
\caption{Averaged classification error rate (\%) for the CIFAR100-C and ImageNet-C benchmarks with the continual setting. The error rates are averaged for the given 15 corruption. † indicates that the result is reported by \cite{dobler2023robust}.}
\label{tab:cifar100c}
\end{table}

\begin{table*}[t!]
\centering
\resizebox{1.0\textwidth}{!}{
\begin{tabular}{@{}clcccccccccccccccc@{}}
\toprule
\multicolumn{2}{c}{TIME}                                       & \multicolumn{15}{c}{t$\longrightarrow$}                                                                                                                                                                                                                          &               \\ \midrule
\multicolumn{2}{c|}{Method}                                    & \rotatebox[origin=c]{75}{Gaussian}      & \rotatebox[origin=c]{75}{shot}          & \rotatebox[origin=c]{75}{impulse}       & \rotatebox[origin=c]{75}{defocus}      & \rotatebox[origin=c]{75}{glass}       & \rotatebox[origin=c]{75}{motion}        & \rotatebox[origin=c]{75}{zoom} & \rotatebox[origin=c]{75}{zoom} & \rotatebox[origin=c]{75}{frost} & \rotatebox[origin=c]{75}{fog} & \rotatebox[origin=c]{75}{brightness} & \rotatebox[origin=c]{75}{contrast} & \rotatebox[origin=c]{75}{elastic} & \rotatebox[origin=c]{75}{pixelate} & \multicolumn{1}{c|}{\rotatebox[origin=c]{75}{jpeg}}          & Mean          \\ \midrule
\multirow{9}{*}{WRN28-10}   & \multicolumn{1}{l|}{Source only†} & 72.3          & 65.7          & 72.9          & 46.9         & 54.3          & 34.8          & 42.0         & 25.1          & 41.3          & 26.0          & 9.3          & 46.7         & 26.6          & 58.5          & \multicolumn{1}{c|}{30.3}          & 43.5          \\
                            & \multicolumn{1}{l|}{BN-1†}        & 28.1          & 26.1          & 36.3          & 12.8         & 35.3          & 14.2          & 12.1         & 17.3          & 17.4          & 15.3          & 8.4          & 12.6         & 23.8          & 19.7          & \multicolumn{1}{c|}{27.3}          & 20.4          \\
                            & \multicolumn{1}{l|}{TENT-cont.†}  & 24.8          & 20.6          & 28.6          & 14.4         & 31.1          & 16.5          & 14.1         & 19.1          & 18.6          & 18.6          & 12.2         & 20.3         & 25.7          & 20.8          & \multicolumn{1}{c|}{24.9}          & 20.7          \\
                            & \multicolumn{1}{l|}{AdaContrast†} & 29.1          & 22.5          & 30.0          & 14.0         & 32.7          & 14.1          & 12.0         & 16.6          & 14.9          & 14.4          & 8.1          & 10.0         & 21.9          & 17.7          & \multicolumn{1}{c|}{20.0}          & 18.5          \\
                            & \multicolumn{1}{l|}{CoTTA†}       & 24.3          & 21.3          & 26.6          & 11.6         & 27.6          & 12.2          & 10.3         & 14.8          & 14.1          & 12.4          & 7.5          & 10.6         & 18.3          & 13.4          & \multicolumn{1}{c|}{17.3}          & 16.2          \\
                            & \multicolumn{1}{l|}{EATA}        & 24.3          & 19.3          & 27.6          & 12.6         & 28.6          & 14.4          & 12.0         & 15.9          & 14.6          & 15.4          & 9.6          & 13.3         & 20.6          & 16.3          & \multicolumn{1}{c|}{21.8}          & 17.8          \\
                            & \multicolumn{1}{l|}{SAR}       &  28.3       &   26.0  &   35.8         &     12.7         &      34.6         &           13.9    &             12.0 &             17.5  &          17.6     &          14.9     &       8.2       &            13.0  &          23.5     &          19.5     &\multicolumn{1}{c|}{27.2}              &  20.3             \\
                            & \multicolumn{1}{l|}{RMT†}         & 21.9          & 18.6          & 24.1 & 10.8         & \textbf{23.6} & 12.0          & 10.4         & 13.0          & 12.4          & 11.4          & 8.3          & 10.1         & \textbf{15.2} & 11.3 & \multicolumn{1}{c|}{14.6}          & 14.5          \\
                            & \multicolumn{1}{l|}{DPLOT (ours)} & \textbf{19.4} & \textbf{16.5} & \textbf{22.5}          & \textbf{10.0} & 23.7          & \textbf{11.1} & \textbf{9.6} & \textbf{12.3} & \textbf{11.9} & \textbf{10.7} & \textbf{7.7} & \textbf{9.8} & 15.5        & \textbf{10.8}          & \multicolumn{1}{c|}{\textbf{13.9}} & \textbf{13.7} \\ \midrule
\multirow{9}{*}{WRN40-2A}   & \multicolumn{1}{l|}{Source only} & 28.8          & 22.9          & 26.2          & 9.5          & 20.6          & 10.6          & 9.3          & 14.2          & 15.3          & 17.5          & 7.6          & 20.9         & 14.8          & 41.3          & \multicolumn{1}{c|}{14.7}          & 18.3          \\
                            & \multicolumn{1}{l|}{BN-1}        & 18.4          & 16.1          & 22.3          & 9.0          & 22.1          & 10.6          & 9.7          & 13.2          & 13.2          & 15.3          & 7.8          & 12.1         & 16.3          & 14.9          & \multicolumn{1}{c|}{17.2}          & 14.5          \\
                            & \multicolumn{1}{l|}{TENT-cont.}  & 15.0          & 12.1          & 16.8          & 9.5          & 18.0          & 11.7          & 9.9          & 11.8          & 11.4          & 13.7          & 9.3          & 11.4         & 16.8          & 13.1          & \multicolumn{1}{c|}{19.8}          & 13.4          \\
                            & \multicolumn{1}{l|}{AdaContrast} & 16.2          & 12.6          & 16.9          & 8.3          & 18.1          & 10.0          & 8.4          & 10.7          & 9.8           & 12.0          & \textbf{7.1}          & 8.4          & 14.2          & 12.1          & \multicolumn{1}{c|}{13.8}          & 11.9          \\
                            & \multicolumn{1}{l|}{CoTTA}       & 15.4          & 13.5          & 16.3          & 9.1          & 17.8          & 10.2          & 8.9          & 11.9          & 11.3          & 14.7          & \textbf{7.1}          & 15.0         & 13.8          & 10.7          & \multicolumn{1}{c|}{13.3}          & 12.6          \\
                            & \multicolumn{1}{l|}{EATA}        & 15.3          & 11.7          & 16.6          & 9.0          & 17.3          & 10.8          & 9.1          & 11.4          & 10.6          & 13.5          & 8.8          & 10.6         & 15.5          & 11.7          & \multicolumn{1}{c|}{16.4}          & 12.5          \\
                            & \multicolumn{1}{l|}{SAR}         &  18.1             &  15.9             &     20.5          &      9.0        &   20.9            &     10.6          &        9.7      &        13.2       &      13.3         &         15.2      &             7.8 &           12.1   &        16.2       &         14.9      & \multicolumn{1}{c|}{17.1}              & 14.3               \\
                            & \multicolumn{1}{l|}{RMT}         & 15.3          & 12.5          & 15.4          & 8.7          & 15.8          & 9.6           & 8.1          & \textbf{9.7}           & 9.6           & 10.4          & 7.2          & 9.9          & \textbf{11.3}          & \textbf{8.8}           & \multicolumn{1}{c|}{\textbf{11.4}}          & 10.9          \\
                            & \multicolumn{1}{l|}{DPLOT (ours)} & \textbf{12.4} & \textbf{10.5} & \textbf{13.9} & \textbf{7.8} & \textbf{14.9} & \textbf{9.1}  & \textbf{8.0} & \textbf{9.7}  & \textbf{9.0}  & \textbf{9.9}  & 7.2 & \textbf{8.3} & 11.6 & \textbf{8.8}  & \multicolumn{1}{c|}{11.7} & \textbf{10.2}  \\ \midrule
\multirow{9}{*}{ResNet-18A} & \multicolumn{1}{l|}{Source only} & 20.2          & 17.5          & 29.3          & 8.8          & 21.7          & 10.5          & 8.7          & 13.5          & 13.5          & 21.6          & 7.2          & 34.9         & 14.3          & 17.1          & \multicolumn{1}{c|}{11.8}          & 16.7          \\
                            & \multicolumn{1}{l|}{BN-1}        & 14.9          & 13.4          & 20.1          & 9.1          & 22.0          & 10.6          & 9.9          & 13.5          & 13.7          & 16.7          & 8.6          & 12.8         & 16.7          & 12.5          & \multicolumn{1}{c|}{15.1}          & 14.0          \\
                            & \multicolumn{1}{l|}{TENT-cont.}  & 13.1          & 11.4          & 17.7          & 9.2          & 19.8          & 12.3          & 10.9         & 13.5          & 12.8          & 16.6          & 10.4         & 11.4         & 16.3          & 12.2          & \multicolumn{1}{c|}{16.2}          & 13.6          \\
                            & \multicolumn{1}{l|}{AdaContrast} & 13.2          & 11.2          & 16.0          & 8.5          & 17.9          & 9.8           & 8.5          & 11.2          & 9.5           & 13.5          & 6.9          & 8.2          & 13.6          & 10.1          & \multicolumn{1}{c|}{10.9}          & 11.3          \\
                            & \multicolumn{1}{l|}{CoTTA}       & 13.6          & 11.9          & 15.7          & 8.6          & 17.2          & 9.3           & 8.5          & 11.3          & 11.2          & 13.9          & 7.4          & 11.0         & 12.7          & 9.6           & \multicolumn{1}{c|}{11.1}          & 11.5          \\
                            & \multicolumn{1}{l|}{EATA}        & 13.0          & 10.9          & 16.1          & 8.5          & 17.1          & 10.0          & 8.7          & 10.6          & 10.2          & 13.9          & 7.9          & 9.5          & 14.5          & 10.5          & \multicolumn{1}{c|}{13.3}          & 11.6          \\
                            & \multicolumn{1}{l|}{SAR}         &    14.9           &       13.4        &         20.0      &      9.1        &      21.3         &         10.6      &        9.9      &       13.5        &        13.7       &        16.7       &        8.6      &         12.8     &        16.7       &        12.5       & \multicolumn{1}{c|}{15.1}              &          13.9     \\
                            & \multicolumn{1}{l|}{RMT}         & 13.3          & 10.9          & 14.9          & 8.4          & 15.1          & 9.5           & 7.9          & 9.5           & 9.6           & 10.2          & 7.5          & 9.0          & 10.9          & 8.5           & \multicolumn{1}{c|}{9.7}           & 10.3          \\
                            & \multicolumn{1}{l|}{DPLOT (ours)} & \textbf{10.3} & \textbf{9.1}  & \textbf{13.2} & \textbf{7.2} & \textbf{14.0} & \textbf{7.8}  & \textbf{6.7} & \textbf{8.1}  & \textbf{7.6}  & \textbf{8.9}  & \textbf{5.8} & \textbf{6.5} & \textbf{9.7}  & \textbf{7.0}  & \multicolumn{1}{c|}{\textbf{8.1}}  & \textbf{8.7}  \\ \bottomrule
\end{tabular}}
\caption{Classification error rate (\%) for the continual CIFAR10-C benchmark; the network is trained on clean CIFAR10 and evaluated on continually given corrupted test data. We evaluate our framework with various models: WRN28-10, WRN40-2A, and ResNet18A. The results are averaged over five runs. Also, the best result is indicated in \textbf{bold}. † indicates that the result is reported by \cite{dobler2023robust}.}
\label{tab:cifar10c}
\end{table*}

\begin{table*}[t]
\centering
\resizebox{0.8\textwidth}{!}{
\begin{tabular}{@{}lccccccc@{}}
\toprule
                                 & \multicolumn{3}{c}{CIFAR10-C}                                                & \multicolumn{2}{c}{CIFAR100-C}                                    & \multicolumn{2}{c}{ImageNet-C}                \\ \midrule
\multicolumn{1}{l|}{Method}      & WRN28-10    & WRN40-2A           & \multicolumn{1}{c|}{ResNet-18A}         & ResNext-29A           & \multicolumn{1}{c|}{WRN40-2A}             & ResNet-50            & ResNet-50A           \\ \midrule
\multicolumn{1}{l|}{Source only} & 24.7 / 43.5† & 10.4 / 18.3        & \multicolumn{1}{c|}{9.7 / 16.7}         & 33.6 / 46.4†          & \multicolumn{1}{c|}{34.7 / 46.7}          & 58.4 / 82.0†          & 44.9 / 67.2          \\
\multicolumn{1}{l|}{BN-1}        & 13.7 / 20.4† & 10.5 / 14.5        & \multicolumn{1}{c|}{10.5 / 14.0}        & 29.9 / 35.4†          & \multicolumn{1}{c|}{33.7 / 39.3}          & 48.3 / 68.6†          & 39.3 / 54.8          \\
\multicolumn{1}{l|}{TENT-cont.}  & 20.4 / 25.1† & 15.0 / 18.2        & \multicolumn{1}{c|}{20.0 / 22.9}        & 74.8 / 75.9†          & \multicolumn{1}{c|}{63.5 / 65.8}          & 46.4 / 58.9†          & 38.5 / 47.0          \\
\multicolumn{1}{l|}{AdaContrast} & 12.1 / 15.8† & 8.9 / 10.9         & \multicolumn{1}{c|}{8.2 / 10.0}         & 33.0 / 35.9†          & \multicolumn{1}{c|}{35.9 / 39.1}          & 66.3 / 72.6†          & 56.7 / 61.5          \\
\multicolumn{1}{l|}{CoTTA}       & 10.9 / 14.2† & 8.6 / 11.3         & \multicolumn{1}{c|}{8.0 / 9.7}          & 26.3 / 28.3†          & \multicolumn{1}{c|}{32.8 / 37.2}          & 38.8 / 43.1†          & 32.0 / 33.8          \\
\multicolumn{1}{l|}{EATA}        & 16.0 / 20.6 & 11.7 / 14.5        & \multicolumn{1}{c|}{11.4 / 13.9}        & 32.0 / 34.4          & \multicolumn{1}{c|}{33.1 / 36.8}          & 40.7 / 49.7          & 36.0 / 41.2          \\
\multicolumn{1}{l|}{SAR}         & 13.6 / 20.3 & 8.7 / 11.4         & \multicolumn{1}{c|}{7.6 / 10.0}         & 28.7 / 31.9          & \multicolumn{1}{c|}{30.7 / 34.5}          & 42.8 / 55.8          & 36.5 / 45.7          \\
\multicolumn{1}{l|}{RMT}         & 9.3 / \textbf{10.4}†  & 7.7 / 8.5          & \multicolumn{1}{c|}{11.7 / 12.5}        & 26.4 / 26.9†          & \multicolumn{1}{c|}{31.1 / 32.1}          & 39.3 / 41.5†          & 33.5 / 34.4          \\
\multicolumn{1}{l|}{DPLOT (ours)} & \textbf{8.8} / \textbf{10.4}  & \textbf{7.2 / 8.0} & \multicolumn{1}{c|}{\textbf{6.3 / 7.0}} & \textbf{23.9 / 25.0} & \multicolumn{1}{c|}{\textbf{27.3 / 28.6}} & \textbf{37.2 / 40.2} & \textbf{30.8 / 31.6} \\ \bottomrule
\end{tabular}}
\caption{Classification error rate (\%) for CIFAR10-C, CIFAR100-C, and ImageNet-C benchmarks with the gradual setting. Error rates are separately reported by averaging over all severity levels and averaging only over the highest severity level 5 (@level 1-5 / @level 5). †~indicates that the result is reported by \cite{dobler2023robust}.}
\label{tab:gradual}
\end{table*}

\subsection{Continual setting benchmark}

First we evaluate our TTA framework on continual setting benchmarks. In Table \ref{tab:cifar100c}, we provide the averaged classification error rates for CIFAR100-C and ImageNet-C benchmarks in the continual setting, considering 15 different corruptions. Our framework outperforms other methods for CIFAR100-C and ImageNet-C benchmarks except when using ResNet-50. In addition, we provide full comparison result of our framework with other TTA frameworks on CIFAR10-C in Table \ref{tab:cifar10c}. Our framework achieves state-of-the-art performances, outperforming the best baseline by 5.5\%, 6.4\%, and 15.5\% in mean error rate when using WRN28-10, WRN40-2A, and ResNet-18A, respectively.

\subsection{Gradual setting benchmark}
In Table \ref{tab:gradual}, we report the average error rate across all severity levels and specfically with respect to level 5 in gradual setting benchmarks. We have following observations. First, TENT suffers from the error accumulation as observed in \cite{niu2022efficient, kirkpatrick2017overcoming} for gradual setting benchmarks due to more frequent updates. For example, TENT has an increased mean error rate compared to continual setting (e.g., 60.9 $\rightarrow$ 74.8 when using ResNext-29A). Second, entropy minimization-based methods (TENT, EATA, and SAR) tend to have degraded performance compared to consistency regularization-based methods (CoTTA and RMT) because the self-generated pseudo-label is more susceptible to error accumulation due to model collapse and forgetting~\cite{niu2022efficient, niu2023towards}. Lastly, our framework consistently outperforms other competitive methods across different architectures and datasets, highlighting the benefit of using domain-specific block selection and paired-view consistency for long-time adaptation against corruptions.  



\section{Discussion}
\label{sec:discussion}

\subsection{Component analysis}

To understand the effect of each component of our framework, we provide the adaptation performance with various configurations in Table \ref{tab:config}. First, we present the performance when using all components: the paired-view consistency, EMA teacher, and the block selection (A). If we remove paired-view consistency using teacher-generated pseudo-labels, the performance significantly drops (B). Furthermore, the performance slightly drops if we do not use ensemble prediction as observed in \cite{dobler2023robust} (C). Finally, it is demonstrated that the performance significantly drops when we update BN layers rather than domain-specific blocks by entropy minimization (D). These results show that all components are necessary for stable long-term adaptation.

\begin{table}[t]
\centering
\resizebox{0.48\textwidth}{!}{
\begin{tabular}{@{}l|ccc@{}}
\toprule
Method                    & CIFAR10-C     & CIFAR100-C    & ImageNet-C    \\ \midrule
DPLOT (A)          & \textbf{8.8 / 10.4}   & \textbf{23.9 / 25.0} & \textbf{30.8 / 31.6} \\
 $-$ Paired-view consistency (B)        & 11.3 / 14.2   & 25.9 / 28.8 & 37.4 / 42.5 \\
 $-$ EMA teacher (C)           & 12.5 / 15.8   & 25.8 / 28.5 & 37.9 / 42.8 \\
 $-$ Block selection (D) & 19.3 / 24.5   & 66.9 / 68.7 & 38.5 / 47.0  \\ \bottomrule
\end{tabular}}
\caption{Classification error rate (@level 5/@level 1-5) for the gradual benchmarks with various configurations. We use WRN28-10, ResNext-29A, and ResNet50A for CIFAR10-C, CIFAR100-C, and ImageNet-C datasets, respectively. Note that, we gradually remove each component from DPLOT, and when none of the components are used (D), the method is equivalent to TENT.}
\label{tab:config}
\end{table}

\subsection{Parameter sensitivity}
In Table~\ref{tab:gamma}, we empirically demonstrate the influence of the hyperparameter $\gamma$ for block selection. As $\gamma$ increases, fewer blocks are selected and vice versa. For instance, only one block is selected for entropy minimization if we set $\gamma$ to 0.999, while all blocks except the lowest one are selected for $\gamma$ of 0.0. We find that small $\gamma$ significantly damages the adaptation performance, but it tends to have robust performance for the range of [0.75, 0.95]. The results demonstrate that as blocks not involve in domain-specific feature extraction are selected to be updated by entropy minimization, the model becomes vulnerable to the model collapse (i.e., predicts all samples to one class \cite{niu2022efficient, niu2023towards}). This vulnerability, caused by modified domain-invariant feature space, can be alleviated by our block selection. Also, updating selected blocks using $\gamma$ in the range of [0.75, 0.95] improves adaptation performance compared to updating the BN weights.

\begin{table}[ht]
\centering
\resizebox{0.36\textwidth}{!}{
\begin{tabular}{@{}l|ccc@{}}
\toprule
Threshold $\gamma$             & CIFAR10-C                            & \multicolumn{1}{l}{CIFAR100-C}        & ImageNet-C                            \\ \midrule
0.999                          & 9.2 / 11.4                           & 24.8 / 26.1                           & 37.8 / 41.2                           \\
0.95                           & 8.8 / 10.7                           & 24.4 / 25.6                           & 37.3 / 40.4                           \\
0.9                            & 8.8 / 10.7                           & 24.3 / 25.5                           & 37.1 / 40.2                           \\
0.75                           & 8.8 / 10.4                           & 23.9 / 25.0                           & 37.2 / 40.2 \\
0.5                            & 11.1 / 12.5                          & 23.8 / 24.5 & 39.1 / 41.2                           \\
0.25                           & 12.8 / 14.5                          & 92.6 / 93.5                           & 99.1 / 99.8                           \\
0.0                            & 80.7 / 81.2                          & 92.7 / 93.5                           & 99.6 / 99.9                           \\ \midrule
BatchNorm & 9.6 / 12.8 & 25.4 / 26.9 & 37.7 / 40.7 \\ \bottomrule
\end{tabular}}
\caption{Classification error rate (@level 5/@level 1-5) for the gradual benchmarks with various thresholds $\gamma$ for block selection. For CIFAR10-C, CIFAR100-C, and ImageNet-C, network architecture of WRN28-10, ResNext-29A, and ResNet50 is used, respectively. BatchNorm indicates that block selection is not used and BN weights are updated, as in previous methods \cite{wang2021tent, niu2022efficient, niu2023towards}.}
\label{tab:gamma}
\end{table}

\begin{figure}[t]
    \centering
    \includegraphics[width=0.49\textwidth]{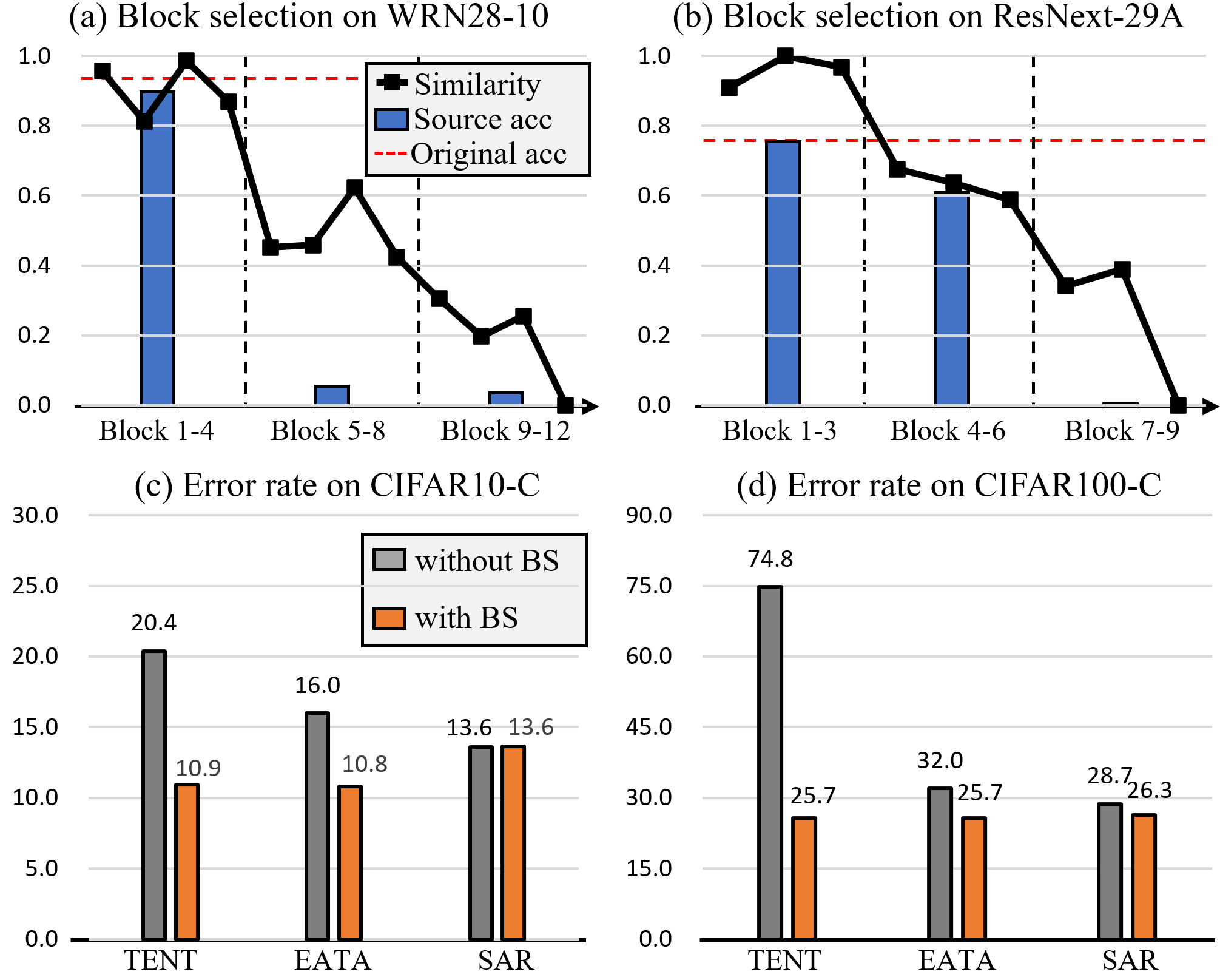}
    \caption{Illustrations of our proposed block selection results (a, b) and classification error rate (@level 1-5) for the gradual setting benchmark using other entropy minimization-based methods with or without block selection (c, d). Additionally, in (a) and (b), the source accuracy after the long-time adaptation (i.e., gradual setting benchmark) with selected blocks is shown in a bar graph.}
    \label{fig:blockselection}
\end{figure}

\subsection{Effect of block selection}


In Figure \ref{fig:blockselection}, we present the results of proposed block selection for WRN28-10 and ResNext-29A for CIFAR10 and CIFAR100, respectively. It is demonstrated that the shallower blocks show high similarity between prototypes before and after entropy minimization, while deeper blocks show lower similarity in (a, b). These findings align with observations in \cite{zhou2021mixstyle, choi2022improving, wang2021domain} that style knowledge (i.e., domain specific feature unrelated to the task) being predominantly captured by shallow blocks. As expected, we find that updating blocks with high similarity does not modify the domain-invariant feature space even after the long-time adaptation (i.e., not forgetting source knowledge (a, b)). Moreover, when applying domain-specific block selection to other methods instead of updating all BN layers (c, d), we find that our block selection methods improves other methods as well. In particular, it reduces the error rate of TENT and EATA by 46.6\% and 32.5\% in CIFAR10-C, respectively. This can be interpreted as updating domain-specific blocks can alleviate error accumulation caused by model collapse, as addressed by \cite{niu2022efficient, niu2023towards}. However, there is no significant improvement in the SAR method. This is because SAR uses a model reset approach, which restores model parameters to their original values; thus, the adaptation performance does not differ significantly.


\begin{figure}[t]
    \centering
    \includegraphics[width=0.47\textwidth]{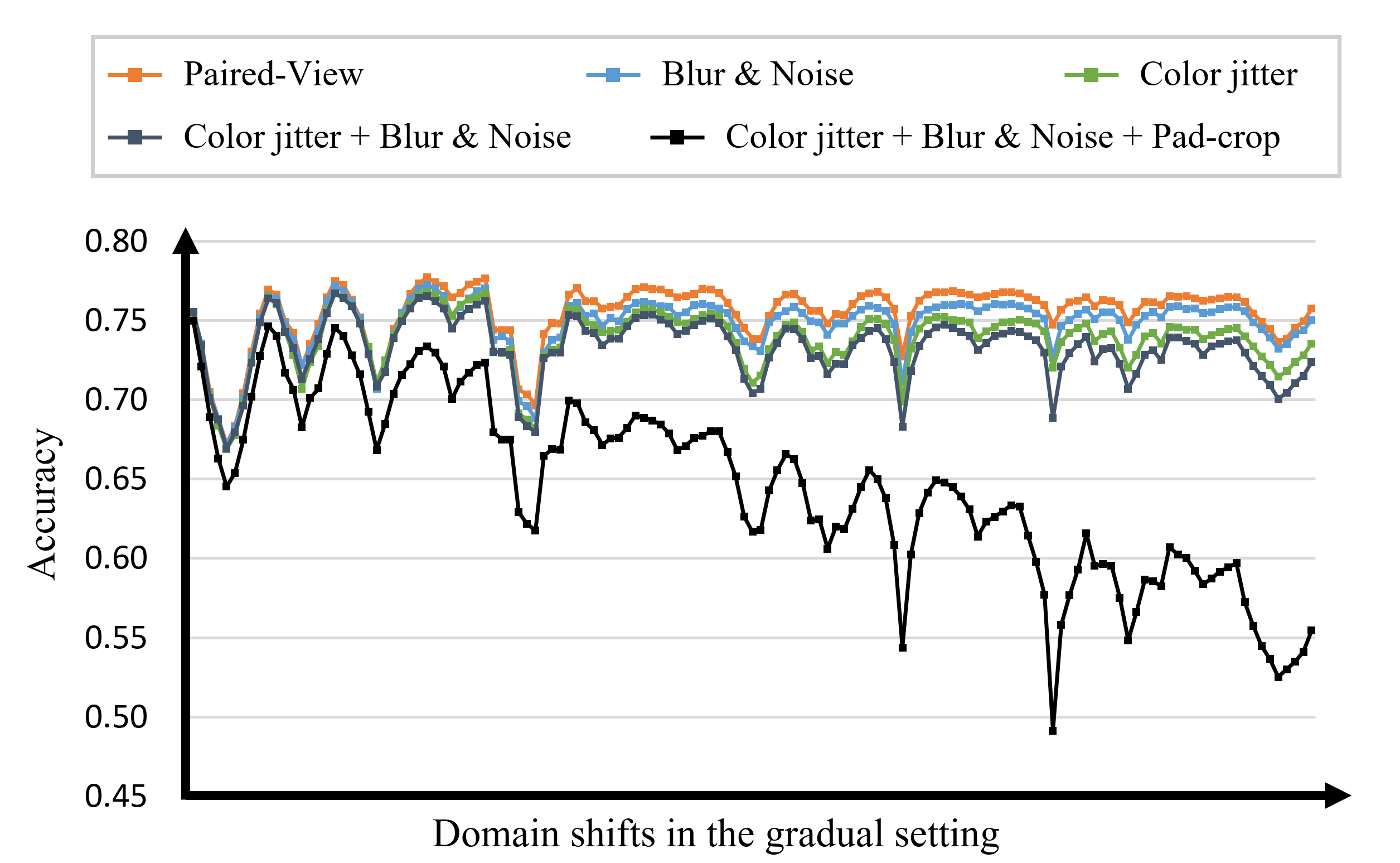}
    \caption{Performance of our framework with various pseudo-label generation setting including our paired-view (orange) and others. We use ResNext-29A for CIFAR100-C.}
    \label{fig:pairedview}
\end{figure}

\subsection{Effect of paired-view consistency}
In CoTTA \cite{wang2022continual}, pseudo-labels are generated by averaging the teacher's predictions for the given 32 augmented images transformed by random augmentation including random pad-crop, color jitter, random affine transform, Gaussian blur, random horizontal flip, and Gaussian noise. In our framework, we generate pseudo-labels by simply averaging the teacher's predictions for two images: original test and horizontally flipped images. This is based on the insight that the simple flip operation does not create domain gap  between the augmented image and the test image, which can reduces the quality of pseudo label. 

In Figure \ref{fig:pairedview}, we compare the adaptation performance of our framework for the gradual setting with different pseudo-label generation methods. Specifically, while entropy minimization on the domain-specific blocks is conducted, the pseudo label is generated by averaging teacher predictions from (i) 2 images with paired-view (orange; ours), (ii) 32 images with Gaussian noise and Gaussian blur (blue), (iii) 32 images with color jitter (green), (iv) 32 images with noise, blur, and color jitter (grey), and (v) 32 images with random pad-crop, affine transform, noise, blur, and color jitter (black; CoTTA). Note that, random horizontal flip with 0.5 probability is included in (ii-v). As expected, the adaptation with our pseudo-label generation outperforms others. It is worth noting that using random pad-crop, which adds a 16-pixel border to the CIFAR image and subsequently performs random cropping to a 32x32 size, significantly decreases the pseudo-label quality due to the potential for objects to be partially cropped.

\subsection{Single sample test-time adaptation}
\label{subsec:ssta}
Since the single prediction for a single input is crucial for some real-time systems, we consider single-sample TTA as investigated in \cite{dobler2023robust}. In the single-sample TTA setting, the last $b$ test samples are stored in a memory buffer. After every~$b$ steps, the model parameters are updated by test-time adaptation methods with a $b$-size batch from the memory. Following \cite{dobler2023robust}, we decrease the learning rate by $\textit{original batch size} / b$ due to the more frequent updates. In this setting, the challenge of the TTA is that error accumulation also increases due to the frequent updates. Table~\ref{tab:singletta} provides the results for single-sample TTA with various buffer sizes $b$. We observed that previous methods suffer from a small batch-size, but our method is relatively strong across various buffer sizes. This demonstrates that our method alleviates error accumulation caused by frequent updates through proper pseudo-label generation.


\begin{table}[]
\centering
\resizebox{0.35\textwidth}{!}{
\begin{tabular}{@{}l|cccc|c@{}}
\toprule
\multirow{2}{*}{Method} & \multicolumn{4}{c|}{Window size}                              & \multirow{2}{*}{Mean} \\ \cmidrule(lr){2-5}
                        & 8             & 16            & 32            & 64            &                       \\ \midrule
Source only             & \multicolumn{4}{c|}{43.5}                                     & 43.5                  \\
BN-1                    & 26.2          & 23.1          & 21.5          & 20.8          & 22.9                  \\
TENT                    & 23.6          & 20.2          & 18.6          & 18.9          & 20.3                  \\
CoTTA                   & 27.4          & 37.5          & 17.1          & 15.0          & 24.3                  \\
EATA                    & 22.9          & 19.2          & 17.6          & 17.9          & 19.4                  \\
RMT                     & 32.3          & 21.8          & 15.8          & 12.4          & 20.6                 \\
DPLOT (ours)             & \textbf{16.4} & \textbf{13.1} & \textbf{11.6} & \textbf{11.3} & \textbf{13.1}         \\ \bottomrule
\end{tabular}}
\caption{Classification rate (@level 1-5) in the gradual setting of the CIFAR10-C benchmark with WRN28-10, using single-sample TTA, while considering different buffer sizes $b$.}
\label{tab:singletta}
\end{table}
\section{Conclusion}
\label{sec:conclusion}
In this work, we propose DPLOT to address proper pseudo-label generation. The proposed framework is based on two components: domain-specific block selection before deployment and paired-view pseudo-labeling. After deployment, we use entropy minimization to update blocks involved in domain-specific feature extraction. Subsequently, we employ paired-view consistency loss, which forces the model to produce the exact prediction of the pseudo-label generated by averaging the teacher's predictions for the test and its corresponding flipped inputs. Extensive experiments demonstrated that our framework outperforms previous competitive methods by a large margin in TTA benchmarks. Also, DPLOT does not modify the network during the training stage, making it easily applicable.




\section*{Acknowledgement}
This work was partially supported by Institute of Information \& communications Technology Planning \& Evaluation (IITP) grant funded by the Korea government (MSIT) (No. 2022-0-00951, Development of Uncertainty Aware Agents Learning by Asking Questions) and by LG Electronics and was collaboratively conducted with the Advanced Robotics Laboratory within the CTO Division of the company.

{
    \small
    \bibliographystyle{ieeenat_fullname}
    \bibliography{main}
}


\end{document}